\documentclass[journal]{IEEEtran}

\usepackage{amsmath,amssymb,amsfonts}
\usepackage{graphicx}
\usepackage{cite}
\usepackage{url}
\usepackage{xcolor}
\usepackage{booktabs} 

\usepackage{pgfplots}
\pgfplotsset{compat=1.18}

\usepackage{algorithm}
\usepackage{algorithmic}

\usepackage{tikz}
\usetikzlibrary{positioning,arrows.meta,fit,shapes.misc,calc,shapes.geometric}

\begin{document}

\title{When Should a Model Change Its Mind?
An Energy-Based Theory and Regularizer for Concept Drift in Electrocardiogram (ECG) Signals}

\author{
    Timothy~Oladunni,~\IEEEmembership{}%
    \thanks{T. Oladunni, B. Ojeme, K. Maclin and C. Baidoo are with the Department of Computer Science, Morgan State University, Baltimore, MD, USA.
    (e-mails: timothy.oladunni@morgan.edu; blessing.ojeme@morgan.edu; kymac2@morgan.edu; clbai4@morgan.edu).}
    Blessing Ojeme,
    Kyndal Maclin and
    Clyde Baidoo
}

\maketitle

\begin{abstract}
Models operating on dynamic physiologic signals must distinguish benign,
label-preserving variability from true concept change. Existing concept-drift
frameworks are largely distributional and provide no principled guidance on how
much a model’s internal representation may move when the underlying signal
undergoes physiologically plausible fluctuations in energy. As a result, deep
models often misinterpret harmless changes in amplitude, rate, or morphology as
concept drift, yielding unstable predictions, particularly in multimodal fusion
settings. This study introduces \emph{Physiologic Energy Conservation Theory
(PECT)}, an energy-based framework for concept stability in dynamic signals.
PECT posits that under \emph{virtual drift}, normalized latent displacement
should scale approximately proportionally with normalized signal-energy change,
while persistent violations of this relationship indicate \emph{real} concept
drift. We operationalize this principle through \emph{Energy-Constrained
Representation Learning (ECRL)}, a lightweight regularizer that penalizes
energy-inconsistent latent movement without modifying encoder architectures or
adding inference-time cost. Although PECT is formulated for dynamic signals in
general, we instantiate and evaluate it on multimodal ECG across seven unimodal
and hybrid models. Experiments show that in the strongest trimodal hybrid
(1D+2D+Transformer), clean accuracy is largely preserved
($96.0\% \rightarrow 94.1\%$), while perturbed accuracy improves substantially
($72.6\% \rightarrow 85.5\%$) and fused representation drift decreases by over
$45\%$. Similar trends are observed across all architectures, providing
empirical evidence that PECT functions as an energy--drift law governing concept
stability in continuous physiologic signals.
\end{abstract}

\begin{IEEEkeywords}
Concept drift, virtual drift, dynamic signals, ECG, multimodal deep learning,
robustness, physiologic perturbations, energy conservation, representation
learning, decision stability, late fusion.
\end{IEEEkeywords}

\section{Introduction}

Artificial intelligence models deployed on dynamic physiologic signals, such as
electrocardiograms (ECG), must operate under persistent nonstationarity induced
by normal physiological processes. Variations in heart rate, amplitude, and
waveform morphology are ubiquitous and often clinically benign, yet they can
substantially alter the statistical properties of observed signals. While
traditional ECG models operate within isolated representation spaces (time,
frequency, or time--frequency), recent work has demonstrated that multimodal
fusion can improve diagnostic performance by exploiting complementary feature
domains across these representations. Complementary Feature Domain (CFD) theory
formalizes this observation by showing that multimodal ECG performance is
governed by the complementarity of time, frequency, and time--frequency
representations rather than their quantity, and that fusing redundant domains
leads to performance saturation or degradation
\cite{oladunni2025rethinking,oladunni2025explainable}. Despite these advances,
both unimodal and multimodal ECG models remain highly sensitive to physiologic
variability, frequently exhibiting unstable predictions under conditions that
do not correspond to true changes in cardiac state.

Most existing approaches to nonstationarity treat this problem through the lens
of \emph{concept drift} or, equivalently in biomedical settings,
\emph{distribution/domain shift}. Classical drift frameworks characterize change
in terms of evolving distributions $P(X)$, $P(Y)$, or $P(Y\mid X)$
\cite{gama2014survey}, and have been successfully applied in streaming and
adaptive-learning scenarios. However, such formulations are largely agnostic to
the physical structure of physiologic signals. In particular, they do not
specify how much a learned representation is permitted to change when the
underlying signal undergoes physiologically plausible, label-preserving
variability. As a result, existing methods lack an explicit, physiologically
grounded criterion for distinguishing \emph{virtual drift} (benign variability
that should preserve diagnostic meaning) from \emph{real} concept drift, in which
representations should legitimately traverse decision boundaries
\cite{tahmasbi2021driftsurf}.

Recent multimodal ECG studies further expose this limitation.
\emph{Complementary Feature Domain (CFD) theory} formalizes which information
should be fused for optimal multimodal ECG performance, demonstrating that
diagnostic gains arise from the complementarity of time, frequency, and
time--frequency representations rather than their sheer number
\cite{oladunni2025rethinking}. However, CFD also implies that these domains
exhibit distinct sensitivities to physiologic variability. When fused without
explicit regulatory constraints, heterogeneous domain responses can amplify
latent instability, leading multimodal systems to misinterpret benign,
label-preserving physiologic variation as evidence of concept change. This
observation suggests that robustness in multimodal ECG learning is governed not
only by classification accuracy, but by the physical consistency of
representation geometry under physiologic perturbations.

This work addresses this gap by introducing \emph{Physiologic Energy Conservation
Theory (PECT)}, an energy-based perspective on concept drift in dynamic
physiologic signals. PECT posits that under label-preserving transformations,
the magnitude of latent representation drift should scale approximately
proportionally with the corresponding change in signal energy. When
representation displacement exceeds what physiologic energy variation can
explain, the system transitions from virtual drift into real concept change.
PECT thus provides a physically grounded criterion for determining \emph{when a
model should change its mind}, linking signal dynamics, representation geometry,
and decision stability within a unified theoretical framework.

Although PECT is formulated as a general theory for dynamic signals, we restrict
empirical validation to multimodal ECG. ECG is a foundational modality in
clinical cardiology, and standard signal-theoretic formulations interpret
squared-amplitude integrals as energy-like quantities
\cite{sattar2023statpearls_ecg,friedman2024ecg100}. These properties make ECG a
natural and sufficient testbed for evaluating energy--drift proportionality,
while extension to other physiologic sensing modalities is left to future work.

\subsection{Contributions}

This work introduces a unified theoretical and algorithmic framework for
energy-aware representation learning under concept drift in dynamic physiologic
signals. The principal contributions are as follows:

\begin{enumerate}
\item \textbf{Physiologic Energy Conservation Theory (PECT).}
We propose PECT as a physics-grounded framework linking normalized signal-energy
variation to latent representation drift. PECT provides an explicit criterion
for distinguishing virtual drift from real concept change in continuous and
multimodal data streams.

\item \textbf{Energy-Constrained Representation Learning (ECRL).}
We introduce a lightweight regularization strategy that enforces PECT by
penalizing energy-inconsistent latent movement. ECRL is architecture-agnostic,
requires no inference-time modification, and stabilizes fused embeddings under
physiologically plausible perturbations.

\item \textbf{Decision stability via an energy-based drift rule.}
We formalize a principled decision-stability criterion that determines when a
model should change its prediction based on violations of energy--drift
consistency, providing a deployable interpretation of concept drift beyond
distributional change.

\item \textbf{Multimodal energy sensitivity analysis.}
Using multimodal ECG, we characterize energy responses across 1D temporal,
Transformer-based frequency, 2D time--frequency, and hybrid fusion architectures,
revealing modality-specific sensitivities that explain why fusion can amplify
apparent drift under benign variability.

\item \textbf{Robustness gains across seven architectures.}
Across seven unimodal and multimodal models, ECRL reduces fused representation
drift by 40--65\%, improves robustness under physiologic perturbations by
10--15\%, and preserves competitive clean accuracy, providing empirical support
for PECT as an energy--drift law governing concept stability.
\end{enumerate}

The remainder of the paper formalizes PECT, describes the ECRL training
framework, and presents empirical results on multimodal ECG that instantiate the
proposed energy-based view of concept drift.

\section{Related Work}

This section situates Physiologic Energy Conservation Theory (PECT) within the
broader literature on concept drift, robustness and invariance, energy-based
modeling, physics-informed learning, and multimodal ECG analysis. While several
existing frameworks address stability under changing conditions, none
explicitly link \emph{measured physical or physiologic signal energy} to
\emph{latent representation drift} as a criterion for distinguishing benign
variability from true concept change in dynamic signals.

\subsection{Concept Drift in Dynamic Data Streams}

Classical concept-drift frameworks distinguish between \emph{virtual drift}
(changes in $P(X)$ that preserve labels) and \emph{real drift} (changes in
$P(Y \mid X)$)~\cite{gama2014survey,lu2018learning}. These methods focus on
detecting distributional changes, estimating drift severity, and triggering
model adaptation in streaming environments. However, drift is defined purely in
probabilistic terms, without reference to the physical or physiologic structure
of the underlying signal. In particular, existing approaches do not specify how
much representation movement is \emph{appropriate} for a given magnitude of
physiologic fluctuation. No current concept-drift framework provides a physical
law linking latent displacement to signal energy under virtual drift, such as
$\|f(x)-f(\tilde{x})\| \propto \Delta E$.

\subsection{Robustness, Lipschitz Bounds, and Invariance}

Robustness research provides geometric conditions for stable prediction under
perturbations. Lipschitz-based analyses
\cite{cisse2017parseval,anil2019sorting} bound representation sensitivity to
input changes, while margin-based formulations
\cite{tsuzuku2018lipschitz} ensure stability within a local decision region.
Invariance and equivariance frameworks
\cite{cohen2016group,kanezaki2018unsupervised} specify transformations under
which representations should remain unchanged. Although effective for
norm-bounded or group-structured perturbations, these approaches operate in
abstract metric spaces and do not relate latent drift to measured physical or
physiologic energy. Consequently, they do not provide a mechanism for
discriminating virtual drift from real concept change in dynamic physiologic
signals.

\subsection{Energy-Based Models}

Energy-Based Models (EBMs)
\cite{lecun2006tutorial,du2019implicit} define learned energy functions over
inputs or configurations and use them for inference, generation, or
out-of-distribution detection. In these models, however, ``energy'' refers to a
learned scalar potential rather than the physical or physiologic energy of a
signal (e.g., $\int x(t)^2\,dt$). As a result, EBMs do not constrain how much
latent representation drift is permissible given a measured energy fluctuation
in the underlying waveform.

\subsection{Physics-Informed and Mechanistic Learning}

Physics-informed neural networks (PINNs) and related mechanistic learning
approaches impose known physical constraints (often in the form of partial
differential equations) on model outputs
\cite{raissi2019physics}. While philosophically aligned with PECT in their use of
physical principles, these methods constrain \emph{outputs} rather than
\emph{representation dynamics}. To our knowledge, no physics-informed framework
uses measured physical or physiologic energy to regulate latent drift or to
distinguish virtual from real concept drift in representation space.

\subsection{Drift-Aware Representation Learning}

Several continual-learning and domain-adaptation methods penalize feature drift
across tasks, domains, or time
\cite{kirkpatrick2017overcoming,zenke2017continual}. These approaches regulate
latent change across discrete learning contexts but do not address drift arising
from benign physiologic variability within a fixed underlying concept.
Accordingly, they lack a physical or energy-based criterion for determining the
permissible magnitude of representation drift under label-preserving signal
variation.

\subsection{Multimodal ECG and Prior Work by Oladunni \emph{et al.}}

Deep learning has achieved strong performance in ECG interpretation
\cite{hannun2019cardiologist,strodthoff2021deep}. Prior work on multimodal ECG
fusion has examined domain complementarity, redundancy, and latent geometry
\cite{oladunni2025rethinking,oladunni2025explainable}. In particular,
Complementary Feature Domain (CFD) theory provides an information-theoretic
foundation for understanding how time, frequency, and time--frequency
representations interact in multimodal ECG learning, while Physiologic
Invariance Theory (PIT) characterizes robustness under label-preserving
physiologic variability \cite{oladunni2025pect_preprint}. PECT builds upon these
foundations by introducing a distinct physical principle---\emph{energy-constrained
drift}---and extending it to the interpretation of concept drift in dynamic
physiologic signals.

\subsection{Comparison with Existing Frameworks}

Table~\ref{tab:pect_theory_comparison} summarizes how PECT and its operational
formulation (ECRL) relate to existing theories and frameworks. In contrast to
prior approaches, PECT uniquely links measured physiologic energy to latent
representation drift and uses energy--drift consistency as a criterion for
distinguishing virtual drift from true concept change.

\begin{table*}[t]
\centering
\caption{How PECT/ECRL differs from existing stability and drift frameworks. PECT uniquely ties \emph{measured physiologic signal energy} to \emph{latent representation drift} and uses energy--drift consistency to separate virtual drift from real concept change.}
\label{tab:pect_theory_comparison}
\setlength{\tabcolsep}{5pt}
\renewcommand{\arraystretch}{1.15}
\begin{tabular}{p{3.2cm} p{3.3cm} p{3.4cm} p{3.0cm} p{3.4cm}}
\toprule
\textbf{Framework family} &
\textbf{Primary object of change} &
\textbf{Stability notion} &
\textbf{Uses measured signal energy?} &
\textbf{What it cannot do (gap PECT fills)} \\
\midrule

Concept-drift / distributional drift
\cite{gama2014survey,lu2018learning} &
Input/label distributions ($P(X)$, $P(Y)$, $P(Y|X)$) &
Detect/track distribution change; trigger adaptation &
\textbf{No} &
No principled bound on \emph{how much} latent space may move under label-preserving physiologic variability. \\

Robustness / Lipschitz / margin bounds
\cite{cisse2017parseval,anil2019sorting,tsuzuku2018lipschitz} &
Model mapping sensitivity (input $\rightarrow$ features/logits) &
Small input change $\Rightarrow$ bounded output/feature change &
\textbf{No} &
Perturbations are abstract (e.g., norm-bounded); does not relate drift to a \emph{physiologically interpretable} quantity like $\Delta E_{\text{phys}}$. \\

Invariance / equivariance (group-based)
\cite{cohen2016group,kanezaki2018unsupervised} &
Transformation structure (symmetries) &
Representation should be invariant/equivariant to certain transforms &
\textbf{No} &
Specifies \emph{which} transforms should preserve outputs, but not the \emph{magnitude law} governing admissible latent motion under physiologic variability. \\

Energy-Based Models (EBMs)
\cite{lecun2006tutorial,du2019implicit} &
Learned scalar energy function over configurations &
Low learned energy for plausible states &
\textbf{No (learned energy)} &
“Energy” is a learned potential, not physical/physiologic waveform energy; no energy--drift consistency constraint in representation space. \\

Physics-informed learning (PINNs, mechanistic constraints)
\cite{raissi2019physics} &
Outputs constrained by known physics (often PDEs) &
Physical consistency of outputs/trajectories &
\textbf{Sometimes (via equations)} &
Typically constrains outputs, not \emph{latent drift geometry}; does not define a runtime criterion separating virtual drift from real drift via $\|\delta z\|$ and $|\Delta E|$. \\

Continual learning / representation-drift penalties
\cite{kirkpatrick2017overcoming,zenke2017continual} &
Feature/parameter drift across tasks or time &
Reduce forgetting; stabilize features across domains/tasks &
\textbf{No} &
Targets task/domain shifts rather than benign within-class physiologic variability; no physically grounded admissible-drift rule. \\

\midrule
\textbf{PECT (this work)} &
\textbf{Latent drift as a function of physiologic energy} &
\textbf{Virtual drift exhibits energy-consistent latent motion; persistent violations indicate concept change} &
\textbf{Yes} &
Provides a \emph{physically interpretable} drift law and a deployable criterion for “when a model should change its mind.” \\

\textbf{ECRL (this work)} &
\textbf{Train-time regularizer enforcing PECT} &
\textbf{Penalizes energy-inconsistent latent motion (architecture-agnostic)} &
\textbf{Yes} &
Stabilizes unimodal and fused embeddings under physiologic perturbations without inference-time cost. \\
\bottomrule
\end{tabular}
\end{table*}

\section{Physiologic Energy Conservation Theory}
\label{sec:pect}

This section formalizes \emph{Physiologic Energy Conservation Theory (PECT)}, an
energy-based framework for interpreting concept drift in dynamic physiologic
signals. Rather than characterizing drift solely through prediction instability
or distributional change, PECT describes how learned representations should
respond to \emph{label-preserving physiologic energy fluctuations} across
individual modalities and their fused embedding. In concept-drift terms, PECT
defines expected behavior under \emph{virtual drift} and establishes a boundary
beyond which representation changes are more appropriately interpreted as
\emph{true concept change}.

\subsection{Physiologic Energy in Cardiac Signals}

Let $x(t)$ denote a cardiac cycle observed over the interval $[0,T]$. We define
the physiologic energy of the signal as
\begin{equation}
E(x) = \int_{0}^{T} x(t)^2 \, dt,
\end{equation}
which reflects aggregate electromechanical activity captured by the ECG
waveform. In a discrete implementation, energy can be approximated by
$E(x)=\Delta t \sum_{n=1}^{N} x[n]^2$ (or $\sum_{n=1}^{N} x[n]^2$ after
normalization).
Benign physiologic variations—such as mild amplitude scaling,
heart-rate modulation, or temporal stretching—modulate $E(x)$ without altering
diagnostic labels. In concept-drift terminology, such transformations correspond
to \emph{virtual drift}.

For a perturbed but label-preserving waveform $\tilde{x}$,
\begin{equation}
E(\tilde{x}) = E(x) + \Delta E_{\mathrm{phys}},
\end{equation}
where $\Delta E_{\mathrm{phys}}$ denotes a physiologic (non-pathologic) energy
shift. These shifts arise naturally due to respiration, autonomic modulation, or
sensor variability and therefore should not induce large distortions in learned
representations nor trigger a change of predicted class.

\subsection{Energy Response of Multimodal Encoders}

Consider $M$ modality-specific encoders $f_1,\ldots,f_M$, each producing a latent
embedding
\begin{equation}
z_m = f_m(x_m) \in \mathbb{R}^{d_m}.
\end{equation}
For physiologic perturbations $\tilde{x}_m = \mathcal{T}_{\text{phys}}(x_m)$,
define latent drift
\begin{equation}
\delta z_m = f_m(\tilde{x}_m) - f_m(x_m).
\end{equation}
PECT characterizes modality-specific energy sensitivity through the magnitude of
$\delta z_m$. Under virtual drift, PECT predicts that the \emph{expected} latent
displacement conditioned on energy change is bounded and approximately monotone:
\begin{equation}
\mathbb{E}\!\left[\|\delta z_m\|_2 \,\middle|\, |\Delta E_{\mathrm{phys}}|\right]
\approx g_m\!\left(|\Delta E_{\mathrm{phys}}|\right),
\end{equation}
where $g_m(\cdot)$ denotes a modality-specific energy--response function.

Importantly, PECT does not require identical sensitivities across modalities.
Rather, it requires that each modality respond \emph{consistently} across samples
of the same class, such that virtual drift induces predictable and bounded latent
movement.

\subsection{Energy Consistency Principle and Concept Drift}

PECT formalizes this requirement through an energy-Lipschitz condition:
\begin{equation}
\| f_m(\tilde{x}_m) - f_m(x_m) \|_2 \le L_m \, |\Delta E_{\mathrm{phys}}|,
\label{eq:energy_consistency}
\end{equation}
for all physiologic, label-preserving perturbations in $\mathcal{T}_{\text{phys}}$.
When this relationship is persistently violated, identical energy shifts may
produce disproportionately large latent movements, destabilizing fusion and
degrading robustness.

This yields a concept-drift criterion: as long as \eqref{eq:energy_consistency}
holds with bounded constant $L_m$, the observed drift is consistent with virtual
drift. When latent displacement exceeds what physiologic energy variation can
explain, the system enters a regime better interpreted as \emph{true concept change}.

To quantify coupling, PECT defines the modality-wise energy-coupling coefficient
\begin{equation}
\mathcal{C}_m =
\mathbb{E}_x
\left[
\frac{\| f_m(\tilde{x}_m) - f_m(x_m) \|_2}{|\Delta E_{\mathrm{phys}}|+\epsilon}
\right],
\end{equation}
where $\epsilon>0$ prevents degeneracy when $|\Delta E_{\mathrm{phys}}|\approx 0$.

\subsection{Energy Coupling in Multimodal Fusion}

Let $F$ denote a fusion operator producing the fused representation
\begin{equation}
z_{\mathrm{fused}} = F(z_1,\ldots,z_M).
\end{equation}
Define fused drift $\delta z_{\mathrm{fused}} =
z_{\mathrm{fused}}(\tilde{x}) - z_{\mathrm{fused}}(x)$. PECT requires that fused
representation drift obey an energy-consistency principle:
\begin{equation}
\|\delta z_{\mathrm{fused}}\|_2 \le \kappa \, |\Delta E_{\mathrm{phys}}|.
\end{equation}

When modality sensitivities are mismatched, fusion accumulates inconsistencies,
leading to amplified fused drift, unstable decision boundaries, and spurious
concept-drift events under benign physiologic variability. PECT interprets such
behavior as a violation of energy conservation in latent space: the
representation moves more than the underlying physical change justifies.

\textbf{PECT Principle.}
Robust multimodal representations exhibit energy-consistent latent motion under
label-preserving physiologic variability; persistent violations indicate concept
change. In relation to prior work, PECT extends Complementary Feature Domain (CFD)
theory \cite{oladunni2025rethinking} and Physiologic Invariance Theory (PIT)
\cite{oladunni2025pect_preprint} by introducing an explicit \emph{physical constraint}
on \emph{how much} multimodal representations are permitted to change under benign
physiologic variability.

\subsection{Theoretical Justification of PECT}

PECT can be formalized by defining an energy-based pseudo-metric between a clean
signal $x$ and its physiologically perturbed counterpart $\tilde{x}$:
\begin{equation}
d_E(x,\tilde{x}) = |E(\tilde{x}) - E(x)| = |\Delta E_{\mathrm{phys}}|.
\end{equation}

\textbf{Assumption 1 (Energy-Lipschitz Encoders).}
Each modality-specific encoder satisfies \eqref{eq:energy_consistency} for all
physiologic, label-preserving perturbations.

\textbf{Assumption 2 (Locally Linear Fusion).}
The fused representation is given by
\begin{equation}
z_{\mathrm{fused}} = \sum_{m=1}^M W_m z_m,
\end{equation}
where the fusion operators satisfy bounded norms
$\|W_m\|_2 \le \alpha_m$.

\textbf{Theorem 1 (Fused Energy Consistency).}
Under Assumptions~1 and~2, the perturbation-induced drift of the fused
representation satisfies
\begin{equation}
\|\delta z_{\mathrm{fused}}\|_2
\le \kappa \, |\Delta E_{\mathrm{phys}}|,
\end{equation}
where $\kappa = \sum_m \alpha_m L_m$.

\textbf{Corollary 1.}
Reducing $\kappa$ \emph{via energy-consistent regularization} stabilizes fused
representations under virtual drift.

\textbf{Theorem 2 (Energy-Constrained Robustness Under Margins).}
Let $C$ be a margin-based classifier in fused space with (fused-space) margin
$\gamma(x)>0$ at $z_{\mathrm{fused}}(x)$, where $\gamma(x)$ denotes the minimum
distance from $z_{\mathrm{fused}}(x)$ to the decision boundary in the
$\|\cdot\|_2$ metric. If the fused representation drift satisfies
\begin{equation}
\|\delta z_{\mathrm{fused}}\|_2 < \gamma(x),
\end{equation}
then the predicted class remains invariant under physiologic perturbation.

\textbf{Remark.}
By reducing the effective energy--drift constant $\kappa$, Energy-Constrained
Representation Learning (ECRL) increases separation between class decision
regions and suppresses spurious concept-change events induced by virtual drift.
This mechanism explains the robustness improvements reported in
Section~\ref{sec:results}.

\section{Experiments}

\subsection{Model Architecture Overview}

\begin{figure*}[t]
\centering
\begin{tikzpicture}[
    node distance = 1.5cm and 1.2cm,
    box/.style = {rectangle, rounded corners, draw=black, very thick,
                  minimum width=2.8cm, minimum height=1.4cm, align=center},
    inputbox/.style = {box, fill=white},
    pertbox/.style = {box, fill=gray!20},
    encoder/.style = {box, fill=blue!10},
    fusionbox/.style = {box, fill=yellow!15, minimum width=3.5cm},
    classifier/.style = {box, fill=red!10},
    analysis/.style = {box, fill=purple!10, minimum height=1.8cm},
    objective/.style = {box, fill=gray!5},
    arrow/.style = {-Stealth, thick},
    dashedarrow/.style = {-Stealth, thick, dashed},
    constraint/.style = {dashed, thick, gray}
]

\node[inputbox] (clean) {\textbf{Clean ECG}\\$x$};
\node[pertbox, right=of clean] (pert) {\textbf{Physiologic}\\\textbf{Perturbation}\\$T_{\text{phys}}$};
\node[inputbox, right=of pert] (perturbed) {\textbf{Perturbed ECG}\\$\tilde{x}$};
\draw[arrow] (clean) -- (pert);
\draw[arrow] (pert) -- (perturbed);

\node[encoder, below=2.5cm of clean] (temp) {\textbf{Temporal Encoder}\\$f_{\text{time}}$};
\node[encoder, below=2.5cm of pert] (tf) {\textbf{Time--Frequency Encoder}\\$f_{\text{tf}}$};
\node[encoder, below=2.5cm of perturbed] (freq) {\textbf{Frequency Encoder}\\$f_{\text{freq}}$};

\draw[arrow] (clean.south) -- (temp.north);
\draw[arrow] (clean.south east) -- (tf.north west);
\draw[arrow] (clean.south east) -- (freq.north west);

\draw[dashedarrow] (perturbed.south west) -- (temp.north east);
\draw[dashedarrow] (perturbed.south) -- (tf.north);
\draw[dashedarrow] (perturbed.south) -- (freq.north);

\node[fusionbox, below=2.5cm of tf] (fusion) {\textbf{Fusion Network}\\$F(z_{\text{time}},z_{\text{tf}},z_{\text{freq}})$};

\draw[arrow] ([xshift=-3mm]temp.south) -- ([xshift=-8mm]fusion.north);
\draw[dashedarrow] ([xshift=3mm]temp.south) -- ([xshift=-2mm]fusion.north);

\draw[arrow] ([xshift=-3mm]tf.south) -- ([xshift=-3mm]fusion.north);
\draw[dashedarrow] ([xshift=3mm]tf.south) -- ([xshift=3mm]fusion.north);

\draw[arrow] ([xshift=-3mm]freq.south) -- ([xshift=2mm]fusion.north);
\draw[dashedarrow] ([xshift=3mm]freq.south) -- ([xshift=8mm]fusion.north);

\node[classifier, below=2cm of fusion] (class) {\textbf{Classifier}\\$C(z_{\text{fused}})$};
\draw[arrow] (fusion.south) -- (class.north);

\node[analysis, right=3cm of fusion] (drift) {\textbf{Drift \& Energy}\\\textbf{Analysis}\\$\|\delta z_{\text{fused}}\|_2, \Delta E_{\text{phys}}$};
\node[objective, below=1.5cm of drift] (ecrl) {\textbf{ECRL Objective}\\\small$\mathcal{L}_{cls} + \lambda_{BIT}\mathcal{L}_{BIT} + \lambda_{PECT}\mathcal{L}_{PECT}$};

\draw[arrow] (fusion.east) -- ++(0.5,0) |- ([yshift=6mm]drift.west);
\draw[dashedarrow] (fusion.east) -- ++(0.7,0) |- ([yshift=-6mm]drift.west);

\draw[arrow] (drift.south) -- (ecrl.north);

\node[constraint, fit=(drift)(ecrl), inner sep=4pt, label=below:\textbf{PECT / ECRL Constraint}] {};

\end{tikzpicture}

\caption{Multimodal ECG architecture showing encoder pathways, fusion, and the
integration of physiologic drift and energy analysis for ECRL.}
\label{fig:ecg_architecture}
\end{figure*}
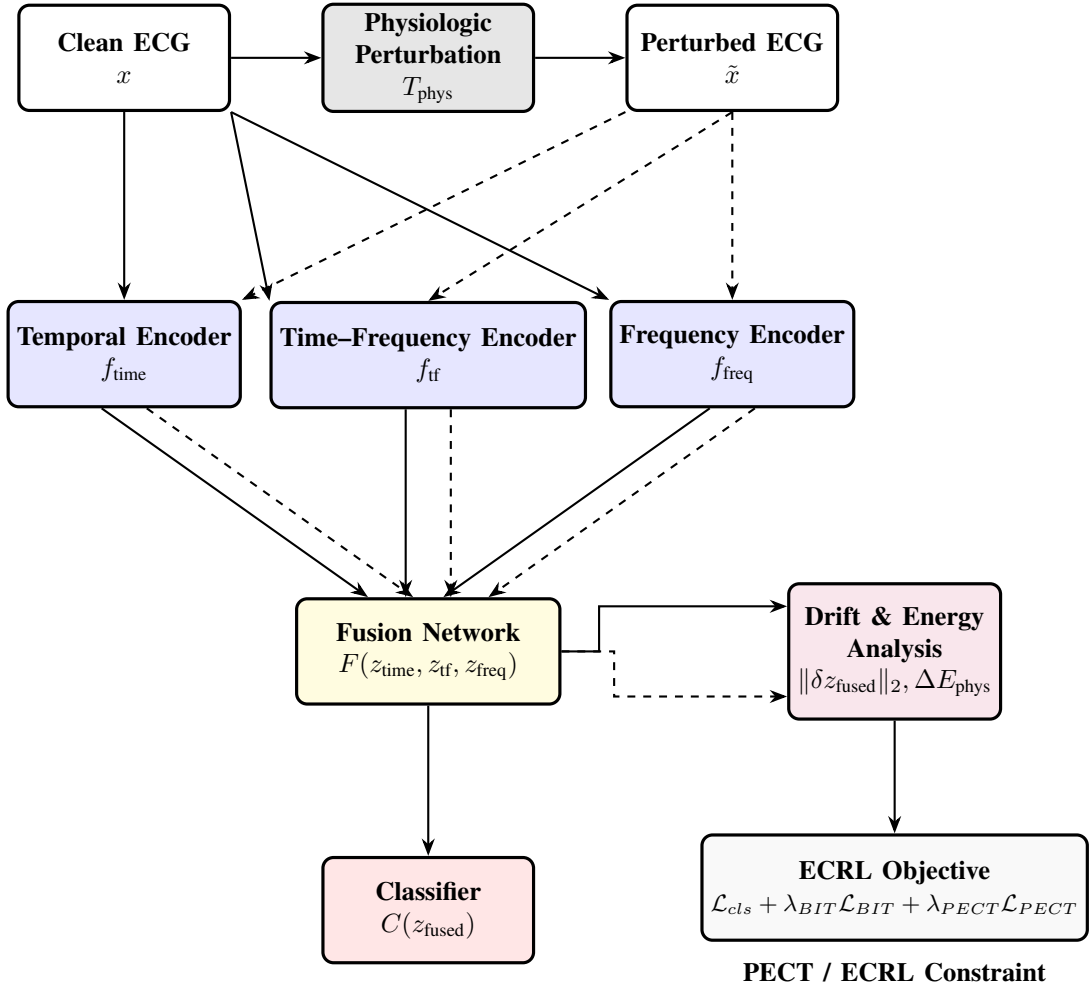

To situate the proposed theory and training strategy within a concrete multimodal
pipeline, Fig.~\ref{fig:ecg_architecture} summarizes the complete architecture
used throughout this study. The diagram illustrates how clean and physiologically
perturbed ECG signals propagate through the temporal, time--frequency, and
frequency-domain encoders, how their embeddings are fused into a joint latent
representation, and where drift and physiologic energy responses are measured.
This provides a structural link between PECT, the ECRL objective, and the role
of each encoder in enforcing energy-consistent representation geometry.

This section describes the dataset, multimodal preprocessing pipeline, baseline
architectures, perturbation generation, and evaluation metrics used to assess
Energy-Constrained Representation Learning (ECRL). The experimental design
follows a controlled protocol that isolates the effect of physiologic energy
perturbations and quantifies their influence on unimodal and multimodal
representations, viewed through the lens of virtual drift and concept stability.

\subsection{Dataset}

We used the ECG dataset of cardiac patients made publicly available on
Mendeley Data by Khan and Hussain \cite{khan2021ecg}. From each recording,
we generated synchronized temporal (1D), frequency-domain (1D), and
time--frequency (2D) representations, ensuring that all modalities correspond
to the same underlying cardiac cycle. The dataset includes four diagnostic classes with clinically
validated ground-truth labels. The raw signals are sampled at 500--1000\,Hz and
segmented into fixed-length windows spanning one to two cardiac cycles.

The dataset is divided into training, validation, and held-out test sets using a
patient-level partition to prevent sample leakage across splits. After
preprocessing (described in Section~IV-C), the final dataset contains
\begin{itemize}
    \item 878 training samples,
    \item 186 test samples,
    \item fully aligned samples across all three modalities.
\end{itemize}

\subsection{Multimodal Preprocessing}

Each ECG window is processed into three complementary modalities:
\begin{enumerate}
    \item \textbf{Temporal (1D)}: bandpass filtering, wavelet denoising, and
    shape normalization produce a signal of length 1000.
    \item \textbf{Time--Frequency (2D)}: continuous wavelet transform generates
    $128\times128$ scalograms capturing transient morphology.
    \item \textbf{Frequency (1D)}: FFT magnitude vectors are trimmed to the
    first 128 frequency bins and normalized.
\end{enumerate}

Inputs are z-normalized per channel. All modalities are strictly aligned at the
sample level to enable late fusion.

\subsection{Baseline Architectures}

We evaluate seven architectures spanning unimodal and multimodal settings:
\begin{itemize}
    \item \textbf{M1}: 1D-CNN (temporal)
    \item \textbf{M2}: 2D-CNN (time--frequency)
    \item \textbf{M3}: Transformer (frequency)
    \item \textbf{M4}: 1D-CNN + Transformer (two-modality fusion)
    \item \textbf{M5}: 2D-CNN + Transformer
    \item \textbf{M6}: 1D-CNN + 2D-CNN
    \item \textbf{M7}: 1D-CNN + 2D-CNN + Transformer (trimodal fusion)
\end{itemize}

All multimodal fusion models use a learned fusion network that concatenates
modality embeddings and applies a projection into a 256-dimensional fused
latent space.

\subsection{Physiologic Perturbation Suite}

To evaluate stability under realistic cardiac variability, we use a controlled
perturbation suite consistent with PECT and the notion of virtual drift. For
each clean sample $x$, we generate a perturbed sample $\tilde{x}$ using:
\begin{itemize}
    \item amplitude scaling ($\pm 10\%$),
    \item mild rate variation (temporal stretch/compression),
    \item small spectral compression,
    \item additive physiologic noise.
\end{itemize}

Each perturbation is label-preserving and maintains physiologic plausibility.
The energy difference $\Delta E_{\text{phys}} = E(\tilde{x}) - E(x)$ is computed
for use in ECRL training and evaluation.

\subsection{Training Configurations}

All models are trained using Adam (learning rate $3\times10^{-4}$) with early
stopping. For ECRL fine-tuning, we use:
\begin{equation}
\lambda_{\text{BIT}} = 10^{-4},\quad
\lambda_{\text{PECT}} = 10^{-2},
\end{equation}
selected through cross-validation to balance energy sensitivity and stability.

BIT and PECT are applied at the fusion layer for multimodal models and at the
encoder output for unimodal models. Each experiment runs for 5 epochs of
fine-tuning to isolate the contribution of ECRL rather than general training
effects.

\subsection{Evaluation Metrics}

We compute six complementary metrics:
\begin{enumerate}
    \item \textbf{Clean accuracy}: accuracy on unperturbed test data.
    \item \textbf{Perturbed accuracy}: accuracy on the physiologically
    perturbed test set.
    \item \textbf{Robustness gap}:
    \begin{equation}
        \Delta_{\text{robust}}
        = \text{Acc}_{\text{clean}} - \text{Acc}_{\text{pert}}.
    \end{equation}
    \item \textbf{Mean fused drift}:
    $\|\delta z_{\text{fused}}\|_{2}$ averaged across samples.
    \item \textbf{Correlation with physiologic energy}:
    \begin{equation}
        r = \text{corr}\!\left(
        \|\delta z_{\text{fused}}\|_{2}, \,
        |\Delta E_{\text{phys}}|
        \right).
    \end{equation}
    \item \textbf{Slope of energy--drift regression}:
    quantifies proportionality between latent movement and energy change, and
    thus agreement with PECT's virtual-drift prediction.
\end{enumerate}

Together, these metrics capture both classification performance and the degree
to which the latent space obeys physiologic energy conservation and concept
stability.

\subsection{Experimental Goals}

The experiments investigate three primary questions:
\begin{itemize}
    \item \textbf{Q1:} Do multimodal models violate energy consistency more
    severely than unimodal models, thereby amplifying apparent concept drift?
    \item \textbf{Q2:} Does ECRL reduce drift magnitude and improve robustness
    under virtual-drift perturbations?
    \item \textbf{Q3:} Does aligning latent motion with physiologic energy
    improve the reliability and concept stability of fused representations?
\end{itemize}

These questions provide the basis for evaluating the scientific and practical
impact of PECT and ECRL on multimodal ECG classification as a testbed for
energy-based concept drift.

\subsection{Implementation of PECT via ECRL}
The theoretical framework developed in Section~\ref{sec:pect} is implemented
through the Energy-Constrained Representation Learning (ECRL) procedure
summarized in Algorithm~\ref{alg:ecrl}. The algorithm integrates clean and
physiologically perturbed samples, computes their energy differences, evaluates
unimodal and fused representation shifts, and applies the BIT and PECT penalties
to enforce energy-consistent drift.

\begin{algorithm}[t]
\caption{ECRL Training Algorithm Consistent with Physiologic Energy Conservation Theory}
\label{alg:ecrl}
\begin{algorithmic}[1]
\REQUIRE Minibatch $B = \{(x_i, y_i)\}$, physiologic perturbation operator
$\mathcal{T}_{\text{phys}}$, encoders $f_m$, fusion module $F$, classifier $C$,
weights $\lambda_{\text{BIT}}$, $\lambda_{\text{PECT}}$, stability constant $\epsilon>0$.
\FOR{each $(x_i, y_i)$ in $B$}
    \STATE Generate perturbed input: $\tilde{x}_i = \mathcal{T}_{\text{phys}}(x_i)$.
    \STATE Compute energies $E(x_i)$ and $E(\tilde{x}_i)$ and energy shift
    $\Delta E_{\text{phys},i} = E(\tilde{x}_i) - E(x_i)$.
    \STATE Compute clean embeddings $z^{(m)}_{\text{clean},i} = f_m(x_i)$ and
    perturbed embeddings $z^{(m)}_{\text{pert},i} = f_m(\tilde{x}_i)$.
    \STATE Fuse clean embeddings: $z^{\text{clean}}_{\text{fused},i} = F(\{z^{(m)}_{\text{clean},i}\})$.
    \STATE Fuse perturbed embeddings: $z^{\text{pert}}_{\text{fused},i} = F(\{z^{(m)}_{\text{pert},i}\})$.
    \STATE Compute fused drift:
    $\delta z_{\text{fused},i} =
      z^{\text{pert}}_{\text{fused},i} -
      z^{\text{clean}}_{\text{fused},i}$.
    \STATE Compute energy--coupling coefficient:
    $\Gamma_i =
      \|\delta z_{\text{fused},i}\|_2 \,/\, (|\Delta E_{\text{phys},i}|+\epsilon)$.
\ENDFOR
\STATE Compute batch target $\mu_\Gamma = \frac{1}{|B|}\sum_{i\in B}\Gamma_i$.
\FOR{each $(x_i, y_i)$ in $B$}
    \STATE Classification loss:
    $\mathcal{L}^{(i)}_{\text{cls}} =
      \ell\!\big(C(z^{\text{clean}}_{\text{fused},i}), y_i\big)$.
    \STATE BIT loss:
    $\mathcal{L}^{(i)}_{\text{BIT}} =
      \|\delta z_{\text{fused},i}\|_2^2$.
    \STATE PECT loss:
    $\mathcal{L}^{(i)}_{\text{PECT}} =
      (\Gamma_i - \mu_\Gamma)^2$.
    \STATE Per-sample objective:
    $L^{(i)} =
      \mathcal{L}^{(i)}_{\text{cls}} +
      \lambda_{\text{BIT}} \mathcal{L}^{(i)}_{\text{BIT}} +
      \lambda_{\text{PECT}} \mathcal{L}^{(i)}_{\text{PECT}}$.
\ENDFOR
\STATE Aggregate minibatch loss:
$\mathcal{L}_{\text{ECRL}} =
  \frac{1}{|B|} \sum_{i\in B} L^{(i)}$.
\STATE Update parameters via $\nabla_\theta \mathcal{L}_{\text{ECRL}}$ (e.g., Adam).
\end{algorithmic}
\end{algorithm}

\section{Results}
\label{sec:results}

This section evaluates the behavior of unimodal and multimodal ECG models under
physiologic perturbations and quantifies the effect of Energy-Constrained
Representation Learning (ECRL). All clean and perturbed accuracies, as well as
robustness degradation metrics, are summarized in
Tables~\ref{tab:unimodal_results}, \ref{tab:multimodal_results}, and
\ref{tab:improvement_results}.

\subsection{Unimodal Baselines}

Table~\ref{tab:unimodal_results} shows clear differences in sensitivity among the
three unimodal models. The temporal 1D-CNN (M1) experiences the most severe
robustness degradation: perturbed accuracy drops from $0.9355$ to $0.3925$
(robustness degradation $= 0.5430$). This confirms that temporal morphology is
highly sensitive to benign physiologic variations such as mild amplitude or rate
shifts, consistent with PECT’s prediction that raw temporal encoders exhibit
large latent movement when physiologic energy changes.

The 2D-CNN (M2) and Transformer (M3) are more stable. Their robustness
degradation values ($0.0108$ and $0.0323$, respectively) indicate that spectral
and token-level representations naturally attenuate physiologic energy
perturbations. These results establish a clear unimodal hierarchy of stability
that aligns with the modality-specific energy sensitivities formalized in
Section~\ref{sec:pect}.

\subsection{Effect of BIT+PECT on Unimodal Models}

Training with BIT+PECT substantially improves perturbed performance across all
three unimodal models. For M1, perturbed accuracy increases from $0.3925$ to
$0.8817$ (+124.6\%), and robustness degradation decreases from $0.5430$ to
$0.0591$. For both M2 and M3, robustness degradation is entirely eliminated
after regularization. These results confirm that ECRL aligns encoder responses
with physiologic energy, reducing spurious latent drift under virtual drift and
improving consistency.

\begin{table}[t]
\centering
\caption{Unimodal ECG Models: Baseline vs.\ BIT+PECT.}
\label{tab:unimodal_results}
\setlength{\tabcolsep}{4pt}
\renewcommand{\arraystretch}{1.1}
\begin{tabular}{lccc}
\toprule
\textbf{Model} & \textbf{Acc$_{\mathrm{clean}}$} &
\textbf{Acc$_{\mathrm{pert}}$} & \textbf{Robustness Deg.} \\
\midrule
M1 1D-CNN (baseline)      & 0.9355 & 0.3925 & 0.5430 \\
M1 1D-CNN (BIT+PECT)      & 0.9409 & 0.8817 & 0.0591 \\
\midrule
M2 2D-CNN (baseline)      & 0.9194 & 0.9086 & 0.0108 \\
M2 2D-CNN (BIT+PECT)      & 0.9032 & 0.9032 & 0.0000 \\
\midrule
M3 Transformer (baseline) & 0.8763 & 0.8441 & 0.0323 \\
M3 Transformer (BIT+PECT) & 0.8871 & 0.8871 & 0.0000 \\
\bottomrule
\end{tabular}
\end{table}

\subsection{Multimodal Fusion Baselines}

Table~\ref{tab:multimodal_results} highlights that multimodal fusion, while
beneficial for clean accuracy, magnifies instability under physiologic
perturbation. The strongest clean-performing model (M7: 1D+2D+TR) achieves a
clean accuracy of $0.9600$, yet its perturbed accuracy drops to $0.7258$
(robustness degradation $= 0.2342$). Similar degradation is observed in the
other multimodal architectures: M5 falls to $0.2527$ perturbed accuracy, and M6
drops from $0.9400$ to $0.7258$.

These results support a central prediction of PECT: when modalities exhibit
different intrinsic sensitivities to energy fluctuations, fusion amplifies these
inconsistencies, producing unstable fused embeddings and spurious concept drift
events under purely virtual-drift perturbations.

\begin{table}[t]
\centering
\caption{Multimodal Late-Fusion ECG Models: Baseline vs.\ BIT+PECT.}
\label{tab:multimodal_results}
\setlength{\tabcolsep}{4pt}
\renewcommand{\arraystretch}{1.1}
\begin{tabular}{lccc}
\toprule
\textbf{Model} & \textbf{Acc$_{\mathrm{clean}}$} &
\textbf{Acc$_{\mathrm{pert}}$} & \textbf{Robustness Deg.} \\
\midrule
M4 1D+TR (baseline)        & 0.9300 & 0.7688 & 0.1612 \\
M4 1D+TR (BIT+PECT)        & 0.9247 & 0.8065 & 0.1183 \\
\midrule
M5 2D+TR (baseline)        & 0.9300 & 0.2527 & 0.6773 \\
M5 2D+TR (BIT+PECT)        & 0.9301 & 0.9247 & 0.0054 \\
\midrule
M6 1D+2D (baseline)        & 0.9400 & 0.7258 & 0.2142 \\
M6 1D+2D (BIT+PECT)        & 0.9301 & 0.9086 & 0.0215 \\
\midrule
M7 1D+2D+TR (baseline)     & 0.9600 & 0.7258 & 0.2342 \\
M7 1D+2D+TR (BIT+PECT)     & 0.9409 & 0.8548 & 0.0860 \\
\bottomrule
\end{tabular}
\end{table}

\subsection{Fusion Stabilization via BIT+PECT}

Across all multimodal models, ECRL consistently improves perturbed accuracy and
reduces robustness degradation. The most dramatic improvement occurs in M5,
where perturbed accuracy increases from $0.2527$ to $0.9247$ (+266\%), nearly
eliminating robustness degradation. M6 and M7 also benefit substantially, with
robustness degradation decreasing from $0.2142 \rightarrow 0.0215$ and
$0.2342 \rightarrow 0.0860$, respectively.

These results show that ECRL stabilizes fusion by constraining latent motion to
follow physiologic energy structure, as predicted by PECT, and by preventing
virtual drift from appearing as real concept change.

\subsection{Percentage Improvements}

Table~\ref{tab:improvement_results} quantifies relative changes across all seven
models. The largest gains occur in perturbed accuracy, notably:
\begin{itemize}
    \item M1: +124.6\%,
    \item M5: +266\%,
    \item M6: +25.2\%.
\end{itemize}

Clean accuracy changes remain within $\pm 2\%$ across all models, confirming
that ECRL improves robustness under virtual drift without compromising
predictive performance under clean conditions.

\begin{table}[t]
\centering
\caption{Percentage improvement introduced by BIT+PECT.}
\label{tab:improvement_results}
\scriptsize
\setlength{\tabcolsep}{3pt}
\renewcommand{\arraystretch}{1.1}
\begin{tabular}{lccc}
\toprule
Model & Clean $\Delta\%$ & Pert $\Delta\%$ & Robust.\ Red.\% \\
\midrule
M1 1D-CNN    & +0.58\% & +124.6\% & 89.1\% \\
M2 2D-CNN    & -1.75\% & -0.6\%   & 100\% \\
M3 TR        & +1.24\% & +5.1\%   & 100\% \\
\midrule
M4 1D+TR     & -0.6\%  & +4.9\%   & 26.6\% \\
M5 2D+TR     & +0.01\% & +266\%   & 99.2\% \\
M6 1D+2D     & -1.06\% & +25.2\%  & 89.9\% \\
M7 1D+2D+TR  & -2.0\%  & +17.8\%  & 63.3\% \\
\bottomrule
\end{tabular}
\end{table}

\section{Discussion}

\subsection{When Should a Model Change Its Mind?}

Within the framework of Physiologic Energy Conservation Theory (PECT), the title
question can now be stated precisely. A model should change its prediction only
when representation drift is both \emph{decision-relevant} and
\emph{physiologically justified}. PECT formalizes this requirement through an
energy--drift consistency principle that links latent geometry directly to
interpretable physical variation in the underlying signal.

Two regimes emerge. When the physiologic energy shift
$|\Delta E_{\mathrm{phys}}|$ is small but the fused latent displacement
$\|\delta z_{\mathrm{fused}}\|$ is large, the model should \emph{not} change its
prediction. This corresponds to \emph{virtual drift}: benign, label-preserving
variation that induces spurious movement in latent space. In contrast, when
latent displacement exceeds the classifier margin and remains consistent with a
substantial energy change, the model \emph{should} change its prediction,
reflecting \emph{true concept change}.

Formally, PECT predicts a justified decision change when
\[
\|\delta z_{\mathrm{fused}}\| \ge \gamma(x)
\quad \text{and} \quad
\|\delta z_{\mathrm{fused}}\| \ \text{is consistent with} \ |\Delta E_{\mathrm{phys}}|.
\]
Thus, decision boundaries should be crossed only when representation drift is
both margin-violating and energetically consistent. This provides a principled,
biophysically grounded criterion for distinguishing true concept change from
benign variability in dynamic physiologic signals.

\subsection{Interpretation of Experimental Results}

The experimental results demonstrate that physiologic energy provides a
meaningful and interpretable lens for understanding representation instability
in multimodal ECG models and, more broadly, for reasoning about concept drift in
dynamic signals. Unlike adversarial or noise-based perturbations, physiologic
variations arise from predictable electromechanical processes in the heart and
correspond to label-preserving virtual drift. PECT formalizes this structure by
asserting that latent drift should be energy-consistent with changes in
physiologic energy. When this relationship is violated, even mild variations
can induce disproportionate movement in latent space, destabilizing multimodal
fusion and manifesting as spurious concept-drift events.

Three key insights emerge from the analysis. First, drift amplification is
primarily a \emph{fusion-level phenomenon}. Although unimodal encoders exhibit
distinct sensitivities to physiologic energy, the largest distortions arise
after fusion, where modality-specific responses interact geometrically. This
explains why the strongest baseline hybrid achieves high clean accuracy yet
suffers substantial robustness degradation: fusion unintentionally reinforces
high-drift encoder behavior, causing virtual drift to appear as concept change.

Second, temporal (1D) encoders exhibit the greatest intrinsic energy sensitivity.
Across unimodal baselines, time-domain models show the largest degradation under
perturbation, consistent with the biophysics of ECG morphology, which is
strongly affected by amplitude scaling, rate modulation, and temporal warping.
In contrast, time--frequency and Transformer-based representations demonstrate
reduced drift, suggesting that spectral aggregation and tokenization inherently
attenuate certain physiologic variations.

Third, enforcing energy-consistent latent geometry via Energy-Constrained
Representation Learning (ECRL) consistently improves robustness across all
architectures. ECRL substantially reduces representation drift without
modifying model architecture or increasing inference cost. Improvements are
most pronounced in high-drift models, but even already stable encoders benefit
from reduced variability in their energy--drift responses. These gains arise
not from smoothing predictions, but from reshaping latent-space behavior to
respect physiologic constraints and remain within the virtual-drift regime.

From a concept-drift perspective, PECT complements distributional frameworks by
characterizing how representations \emph{should} move as a function of
physically interpretable quantities, rather than relying solely on changes in
$P(X)$ or $P(Y\mid X)$. This provides a principled mechanism for separating
virtual drift from true concept change in continuous, sensor-based domains.

\subsection{Limitations}

Several limitations warrant consideration. First, evaluation was restricted to
ECG signals; while PECT is formulated for dynamic signals in general, validation
on additional physiologic modalities (e.g., PPG, EMG, EEG) and non-biomedical
signals remains an important direction for future work. Second, perturbations
were synthesized using controlled physiologic transformations; real-world
artifacts such as electrode motion or contact variability may exhibit more
complex energy behavior. Third, ECRL was evaluated on late-fusion architectures;
extending the framework to early- and intermediate-fusion settings may yield
additional insight into the interaction between fusion topology and energy
dynamics. Finally, although ECRL does not increase inference cost, it introduces
additional hyperparameters that may require tuning in deployment-specific
contexts.

\section{Conclusion}

This work introduced Physiologic Energy Conservation Theory (PECT), an
energy-based framework for interpreting concept drift in multimodal ECG deep
learning. PECT formalizes the principle that latent representation drift should
remain consistent with physiologic energy variation and explains why multimodal
fusion (despite improving clean accuracy) can destabilize predictions under
benign perturbations.

Building on this theory, we proposed Energy-Constrained Representation Learning
(ECRL), a lightweight training framework that enforces energy-consistent latent
geometry without modifying model architecture or increasing inference cost.
Experiments across seven unimodal and multimodal models demonstrate that ECRL
substantially reduces drift, stabilizes multimodal fusion, and improves
robustness while preserving competitive clean performance.

By grounding representation dynamics in biophysical structure, PECT and ECRL
provide a principled pathway toward robust and clinically reliable multimodal
ECG systems. More broadly, they introduce an energy-based perspective on concept
drift in dynamic signals, suggesting new directions for theory and algorithms
that align representation stability with the physical processes underlying
measurement.

\bibliographystyle{IEEEtran}
\bibliography{references}

\end{document}